\documentclass[sigconf]{acmart}

\AtBeginDocument{%
  }

\usepackage{makecell}
\usepackage{subcaption}

\setcopyright{acmlicensed}
\copyrightyear{2018}
\acmYear{2018}
\acmDOI{XXXXXXX.XXXXXXX}
\acmISBN{978-1-4503-XXXX-X/2018/06}

\begin{document}

%

\title{Leveraging Large Language Models and Traditional Machine Learning Ensembles for ADHD Detection from Narrative Transcripts}

\author{Yuxin Zhu}
\authornote{Both authors contributed equally to this research.}
\email{yuxin.zhu@emory.edu}
\author{Yuting Guo}
\authornotemark[1]
\email{yuting.guo@emory.edu}
\affiliation{%
  \institution{Emory University}
  \city{Atlanta}
  \state{Georgia}
  \country{USA}
}

\author{Noah Marchuck}
\email{noah.marchuck@emory.edu}
\affiliation{%
  \institution{Emory University}
  \city{Atlanta}
  \state{Georgia}
  \country{USA}
}

\author{Abeed Sarker}
\email{abeed.sarker@emory.edu}
\affiliation{%
  \institution{Emory University}
  \city{Atlanta}
  \state{Georgia}
  \country{USA}
}

\author{Yun Wang}
\email{yun.wang2@emory.edu}
\affiliation{%
  \institution{Emory University}
  \city{Atlanta}
  \state{Georgia}
  \country{USA}
}

\renewcommand{\shortauthors}{Zhu et al.}

\begin{abstract}
%The integration of large language models (LLMs) with traditional machine learning methods in psychiatric applications remains largely unexplored. In this paper, we present a novel ensemble-based framework for classifying Attention-Deficit/Hyperactivity Disorder (ADHD) from narrative transcripts. We formulate the task as binary classification and develop three models: LLaMA3 (an open-source LLM), RoBERTa (a fine-tuned transformer-based model), and a Support Vector Machine (SVM). These models are combined using a majority voting scheme. Experimental results show that the ensemble approach achieves the highest performance, with an F$_1$ score of 0.71 and recall of 0.91, outperforming all individual models. Our findings highlight the promise of hybrid modeling strategies for enhancing clinical text classification in the psychological domain.
Despite rapid advances in large language models (LLMs), their integration with traditional supervised machine learning (ML) techniques that have proven applicability to medical data remains underexplored. This is particularly true for psychiatric applications, where narrative data often exhibit nuanced linguistic and contextual complexity, and can benefit from the combination of multiple models with differing characteristics. Prior research in clinical natural language processing (NLP) has primarily focused on fine-tuning transformer-based models or building domain-specific LLMs, but hybrid approaches that combine deep contextual models with classical ML algorithms have been underexplored. In this study, we introduce an ensemble framework for automatically classifying Attention-Deficit/Hyperactivity Disorder (ADHD) diagnosis (binary) using narrative transcripts. Our approach integrates three complementary models: LLaMA3, an open-source LLM that captures long-range semantic structure; RoBERTa, a pre-trained transformer model fine-tuned on labeled clinical narratives; and a Support Vector Machine (SVM) classifier trained using TF-IDF-based lexical features. These models are aggregated through a majority voting mechanism to enhance predictive robustness. The dataset includes 441 instances, including 352 for training and 89 for validation. Empirical results show that the ensemble outperforms individual models, achieving an F$_1$ score of 0.71 (95\% CI: [0.60-0.80]). Compared to the best-performing individual model (SVM), the ensemble improved recall while maintaining competitive precision. This indicates the strong sensitivity of the ensemble in identifying ADHD-related linguistic cues. These findings demonstrate the promise of hybrid architectures that leverage the semantic richness of LLMs alongside the interpretability and pattern recognition capabilities of traditional supervised ML, offering a new direction for robust and generalizable psychiatric text classification.

\end{abstract}


\keywords{Large Language Models, Text Classification, Attention-Deficit{\slash}Hyperactivity Disorder, Machine Learning, Electronic Health Records, Natural Language Processing}

\maketitle

\section{Introduction}
 
Attention-Deficit/Hyperactivity Disorder (ADHD) is a highly heterogeneous neurodevelopmental condition, presenting diverse etiologies, clinical profiles, and comorbidities \cite{PMID:28527021}. No single risk factor or biomarker conclusively accounts for ADHD’s onset; instead, multiple genetic, environmental, and neurodevelopmental factors interplay, leading to varied symptom manifestations in different individuals. Such heterogeneity poses a fundamental challenge for diagnosis. Clinicians rely on behavioral assessments and patient history, but ADHD symptoms (inattention, hyperactivity, impulsivity) often overlap with other disorders (mood, anxiety, learning disorders), making differential diagnosis difficult \cite{adhd1}. There is no single test or “gold-standard” measure to diagnose ADHD in practice---diagnosis typically requires extensive interviews, standardized rating scales from parents and teachers, and the exclusion of alternative explanations \cite{adhd1}. Such comprehensive evaluations are time-consuming and subjective, leading to variability in diagnostic outcomes. In particular, subjectivity and informant bias can affect clinical assessments; parent and teacher ratings may differ widely, and self-reports are prone to error. This subjectivity can yield inconsistent diagnoses, highlighting the need for more objective, reproducible diagnostic aids.

Neuroimaging has been explored as an objective route to ADHD diagnosis, but with limited success. Over two decades of MRI-based studies have revealed some group-level differences in brain structure and function, yet no reliable imaging biomarker has emerged for individual diagnosis. The neuroimaging literature is voluminous and often inconclusive. Meta-analyses, for instance, of ADHD MRI studies frequently find inconsistent or non-overlapping results across samples \cite{yu2023meta, Parlatini}. Variation in normal brain development, ADHD’s own heterogeneity, and small effect sizes mean that MRI findings have not translated into clinically useful tests \cite{PULINI2019108}. In summary, traditional diagnostic methods---clinical evaluation and neuroimaging---are limited by subjectivity, heterogeneity of presentations, and lack of definitive biomarkers. This creates a strong rationale for exploring other computational techniques that can enhance diagnostic accuracy by integrating multi-modal data and reducing human bias.

Recent years have seen growing interest in applying machine learning and data-driven methods to improve ADHD identification and prognosis. Conventional machine learning on clinical datasets has achieved moderate success; for example, decision tree classifiers on neuropsychological and demographic data attained about 75.03\% accuracy in distinguishing ADHD cases from non-ADHD controls \cite{10.3389/fpsyt.2023.1164433}, providing interpretable rules to assist clinicians. Similarly, support vector machines and random forests have been applied to behavioral questionnaires and social media text. one study using a random forest on ADHD-related Reddit posts achieved 81\% accuracy in identifying users with ADHD traits \cite{Alsharif}. These approaches illustrate the promise of algorithmic classifiers but often remain dataset-specific and can struggle with generalizability due to limited feature scopes.

\begin{figure*}[htbp!]
  \centering
  \includegraphics[width=1.5\columnwidth]{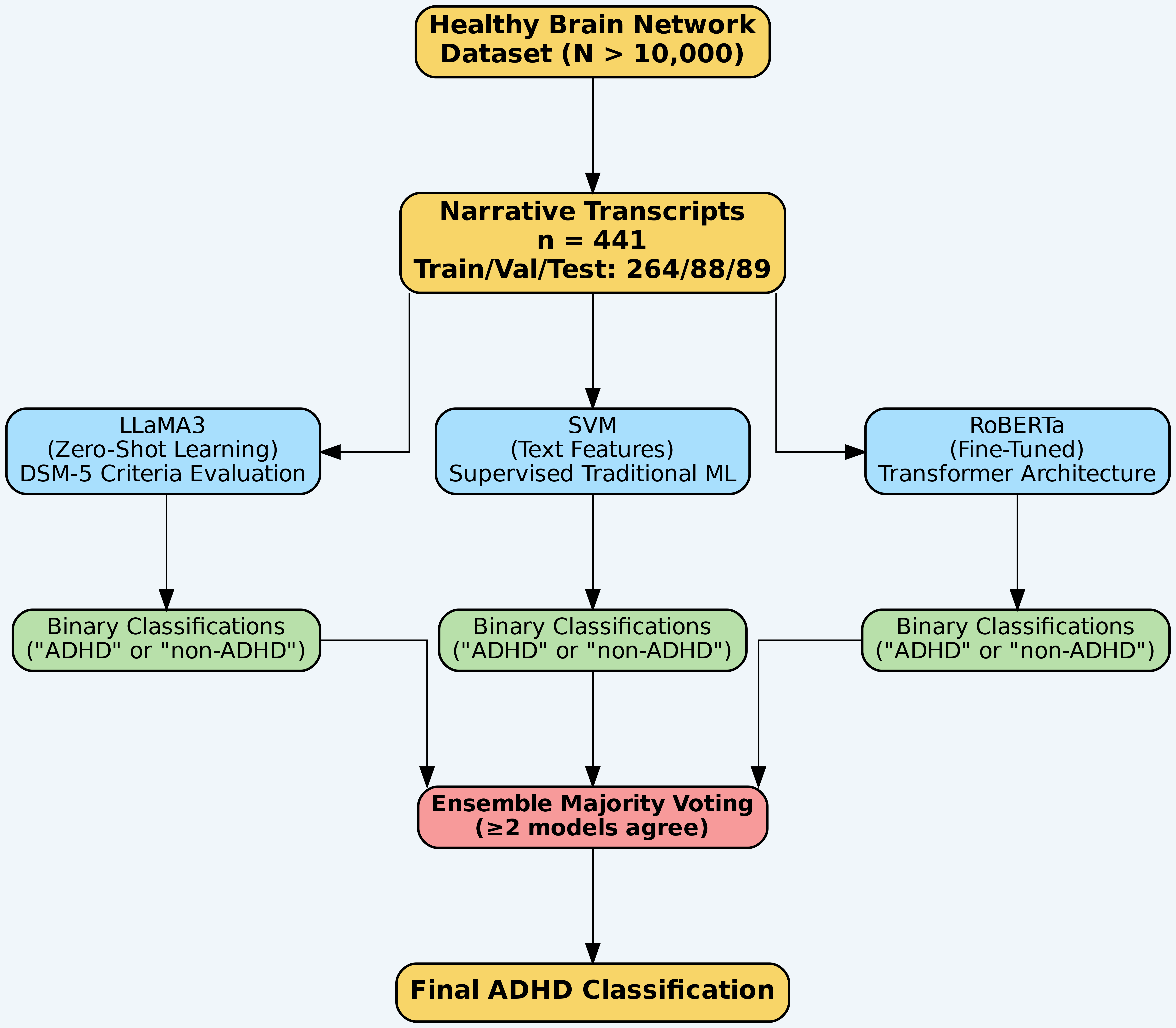}
  \caption{End-to-end ensemble-based narrative classification pipeline. Narrative transcripts elicited immediately after fMRI scanning (“post-scan interviews”) undergo preprocessing, including tokenization, TF-IDF vectorization, engineered feature extraction (e.g., response length, question length), and isolation of interviewee texts, before being fed into three complementary classifiers: (1) LLaMA3 via optimized prompt engineering; (2) a fine-tuned RoBERTa transformer; and (3) a support vector machine leveraging both TF-IDF lexical features and additional engineered metrics. Each model independently generates an ADHD vs. non-ADHD prediction, and these are combined under a majority-voting rule (narratives are labeled ADHD if at least two models concur) to produce the final diagnostic classification. Directed arrows denote the flow of data through each module, illustrating the modular architecture of the ensemble framework.}
    \label{fig:eb}
\end{figure*}

Natural language processing (NLP) offers another avenue, especially given that clinical text (doctor’s notes, psychological reports) and patient self-descriptions contain rich information about ADHD symptoms. NLP techniques have previously been used to detect signs of ADHD in unstructured text. For instance, Malvika et al. applied a transformer-based model (BioClinicalBERT) to electronic health records and achieved an F$_1$ score of 0.78 in identifying ADHD cases \cite{10.1093/jamia/ocae001}, demonstrating the utility of textual data in diagnosis. Likewise, analysis of social media and patient narratives has proven fruitful: fine-tuned transformer models like RoBERTa \cite{Liu2019a} have reached $\sim$76\% accuracy in classifying ADHD-related posts from online forums \cite{lee2024detectingproxypotentialcomorbid}, and SVM classifiers using linguistic and memory-related features attained F$_1$ $\approx 0.77$ in distinguishing ADHD in personal essays \cite{cafiero-etal-2024-harnessing}. These advances demonstrate that textual markers of ADHD, such as patterns of language usage, and descriptions of attention difficulties, can be learned by machine learning models, providing a scalable complement to traditional assessments. Notably, researchers are also prioritizing explainability in these models. Kim et al. developed an explainability-enhanced classifier on psychological test reports, which not only achieved high accuracy ($\sim$92\%) in separating ADHD from intellectual disability, but could also highlight textual evidence (n-gram features) to justify its predictions \cite{Korean}. This ability to provide evidence-based insights is crucial for physician trust in AI-assisted diagnosis.

\begin{figure*}[htbp!]
  \centering
  \includegraphics[width=1.2\columnwidth]{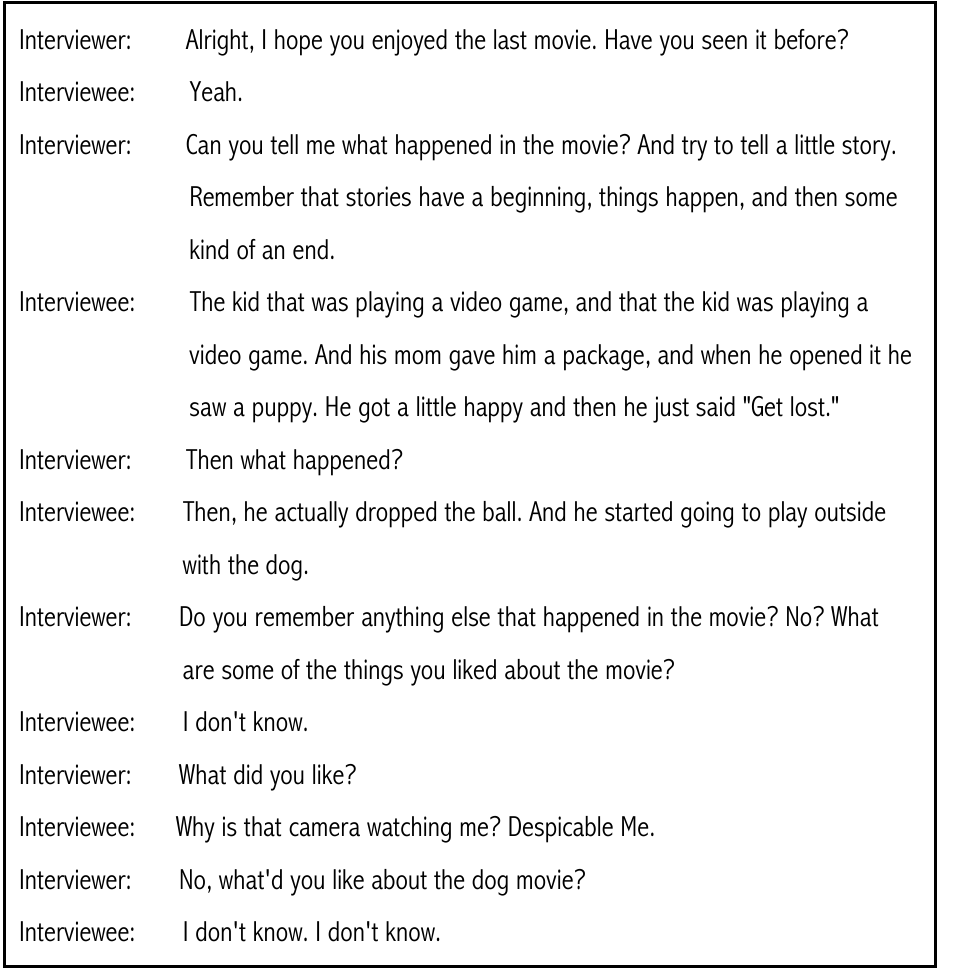}
  \caption{Example of narrative data from a post-scan interview illustrating ADHD-related response patterns. The participant’s answers are fragmented and lack coherence, with shifts in focus and expressions of confusion (e.g., "I don't know"). These behaviors—difficulty staying on topic and providing detailed responses—are characteristic of ADHD and serve as key indicators in the classification process.}
    \label{fig:excerpt}
\end{figure*}

In parallel, large language models (LLMs), such as GPT series \cite{Brown2020} developed by OpenAI and LLaMA \cite{llama3modelcard} developed by Meta, have demonstrated extraordinary capabilities in understanding and generating human-like text. These models, trained on massive corpora, have been shown to exhibit cognitive-like competencies. They can perform tasks requiring reasoning, memory, and attention---key domains affected in ADHD. Berrezueta-Guzman et al. explored the use of ChatGPT (GPT-3.5) as a conversational agent to support ADHD therapy, finding that experts rated it highly in empathy and adaptability during simulated therapy sessions. Their study highlights that LLMs can capture subtle aspects of communication and patient interaction, reinforcing the idea that such models understand context relevant to ADHD \cite{future}. Another line of research evaluates LLMs on mental health prediction tasks via zero-shot or few-shot prompts. Xu et al. evaluated models like GPT-4, Alpaca, and FLAN-T5 on tasks including detecting ADHD from text, and found zero-shot LLM performance to be promising yet below specialized models. Notably, after fine tuning these LLMs on mental health data, the performance jumped significantly---their fine-tuned “Mental-Alpaca” outperformed even much larger base models by over 10\% in balanced accuracy \cite{Xu_2024}. This reveals that while LLMs have general knowledge (and potentially the ability to recognize ADHD-related patterns learned from text during pretraining), they may need domain-specific tuning to reach diagnostic accuracy. Fine-tuning endows them with jargon understanding and emphasis on the subtle linguistic cues of ADHD, whereas in a zero-shot setting, they might miss context-specific details. In contrast, zero-shot LLMs bring the advantage of flexibility---they can be deployed without task-specific training data, an appealing trait when labeled data is scarce or costly. Beyond direct diagnosis, LLMs are being integrated with healthcare data pipelines. For example, ensemble approaches with multiple LLMs have been used to improve medical question-answering systems. Xiao et al. introduced an LLM ensemble that uses a weighted majority voting among different LLMs to answer medical queries, achieving higher accuracy than any single model  by reducing variance and bias \cite{Yang2023.12.21.23300380}. 

In the broader mental health domain, the integration of LLMs with more traditional analytical models, like supervised machine learning classifiers, remains a relatively open frontier. Studies have repeatedly demonstrated that in the presence of annotated data for training, supervised models outperform zero- or few-shot LLMs relying on in-context learning \cite{sarker2024nlp}. While generative LLMs like LLaMA3 have excellent capabilities in understanding nuances of complex, unstructured texts, in-context learning strategies have limitations on how many labeled examples can be provided to the LLM to guide its decision making. This is because the context window sizes of LLMs are limited (e.g., LLaMA3 70B has a context window size of 8000 tokens). At the same time, the large number of parameters in LLMs require very large annotated datasets for meaningful fine-tuning. Thus, traditional supervised machine learning models often have a relative advantage over LLMs when annotated data is available for training. There is, however, little past work attempting to combine the complementary capabilities of LLMs and traditional machine learning models.

In this paper, we describe an ensemble-based ADHD classification model to address this gap. Specifically, we model the ADHD diagnosis as a binary classification task and attempt to solve it by blending state-of-the-art NLP---LLaMA3\footnote{We used the LLaMA3-70B model, which we refer to as LLaMA3 for convenience.} and RoBERTa, with established machine learning techniques---Support Vector Machine (SVM). This ensemble is, to our knowledge, the first effort to incorporate an LLM with a fine-tuned transformer-based model and a traditional machine learning model for a psychiatric diagnosis task. The specific contributions of this paper are as follows:
\begin{itemize}
    \item We designed and refined a prompt that specifies the characteristics to consider when deciding if a transcript should be labeled as ADHD or not.
    \item We developed an LLM (LLaMA3-70B), a transformer-based model (RoBERTa) with supervised learning, and a traditional machine learning model (SVM) for the automatic classification of ADHD cases based on the narrative transcripts.
    \item We developed an an effective ensemble classification framework that combined the individual predictions under the majority voting scheme.
    \item We conducted an analysis of the performance and errors for the classifiers which can provide insights for potential future research directions.
\end{itemize}

\section{Materials and Methods}

\subsection{Data Collection and Narrative Elicitation}

Our work utilized a subset of the Healthy Brain Network (HBN) dataset provided by the Child Mind Institute. Following a Data Usage Agreement with Child Mind Institute, we retrieved the verbatim transcripts and clinical diagnosis data for ADHD. The latter is the final diagnosis given by the clinician after administering the KSADS and considering other data and interactions provided as part of participation into the HBN study. Each of the 441 youths (ages 5–21) in our dataset watched an emotionally evocative short animated film (“The Present”) during functional MRI scanning; immediately upon exiting the scanner, they completed a structured post-scan interview. A set of open-ended questions was used to elicit narrative recall (e.g., describing the sequence of events), emotional interpretation (e.g., identifying characters’ feelings), and perspective-taking (e.g., explaining characters’ motivations). These narrative responses formed the transcripts analyzed by our classification models. We chose the transcript data as our target model input because naturalistic narrative language may serve as an alternative, complementary data source that can ecologically assess ADHD. There is growing evidence that children with ADHD exhibit subtle but systematic differences in storytelling---for example, producing narratives that are less coherent, more error-prone, and less richly detailed than those of their peers \cite{Ida}. In the cases of transcripts from participants with ADHD, responses often lacked direct answers to the interviewer's questions or involved redirecting the conversation by asking questions in return. As shown in the interview excerpt (\ref{fig:excerpt}) below, when asked to describe the events of a movie, the participant provided a fragmented and disjointed recount. Additionally, the participant frequently expressed confusion or disinterest when asked to elaborate, such as when responding 'I don't know' or shifting the focus of the conversation back to the interviewer. By analyzing such patterns across the entire dataset, our study aims to develop a reliable approach for identifying ADHD from narrative transcripts using three classification models and their ensemble model. Each model approach brings a unique inductive bias to the task of ADHD detection, which is detailed in the following section. The comprehensive experiment pipeline is shown in Figure \ref{fig:eb}.

\begin{table*}[htbp!]
  \caption{The statistics for the training, development, and test sets.}
  \label{tab:freq}
  \begin{tabular}{cccc}
    \toprule
    \textbf{Dataset}& \textbf{Size} & \textbf{ADHD} & \textbf{non-ADHD}\\
    \midrule
    Train & 264 & 134 (50.76\%) & 130 (49.24\%)\\
    Dev & 88 & 45 (51.14\%) & 43 (48.86\%)\\
    Test & 89 & 45 (50.56\%) & 44 (49.44\%)\\
  \bottomrule
\end{tabular}
\end{table*}

\subsection{Classification Models}

We split the dataset into a training, development, and test set with a ratio of 60/20/20. All classification models were trained using the same training and development set and evaluated on the same test set. By holding the input data constant across models, we ensure that any improvement from combining models is due to their complementary modeling strengths rather than differences in data. Texts were preprocessed by separating interviewer and interviewee texts. For the LLM, the training set was used for testing and optimizing prompts prior to execution on the test set. The dataset maintained a balanced distribution between participants diagnosed with ADHD and non-ADHD (stratified sampling), thus ensuring consistency in class representation throughout all experimental stages. The detailed data statistics are shown in Table \ref{tab:class}.

\subsubsection{Large Language Model: LLaMA3}

\begin{figure*}[ht]
\includegraphics[scale=0.4]{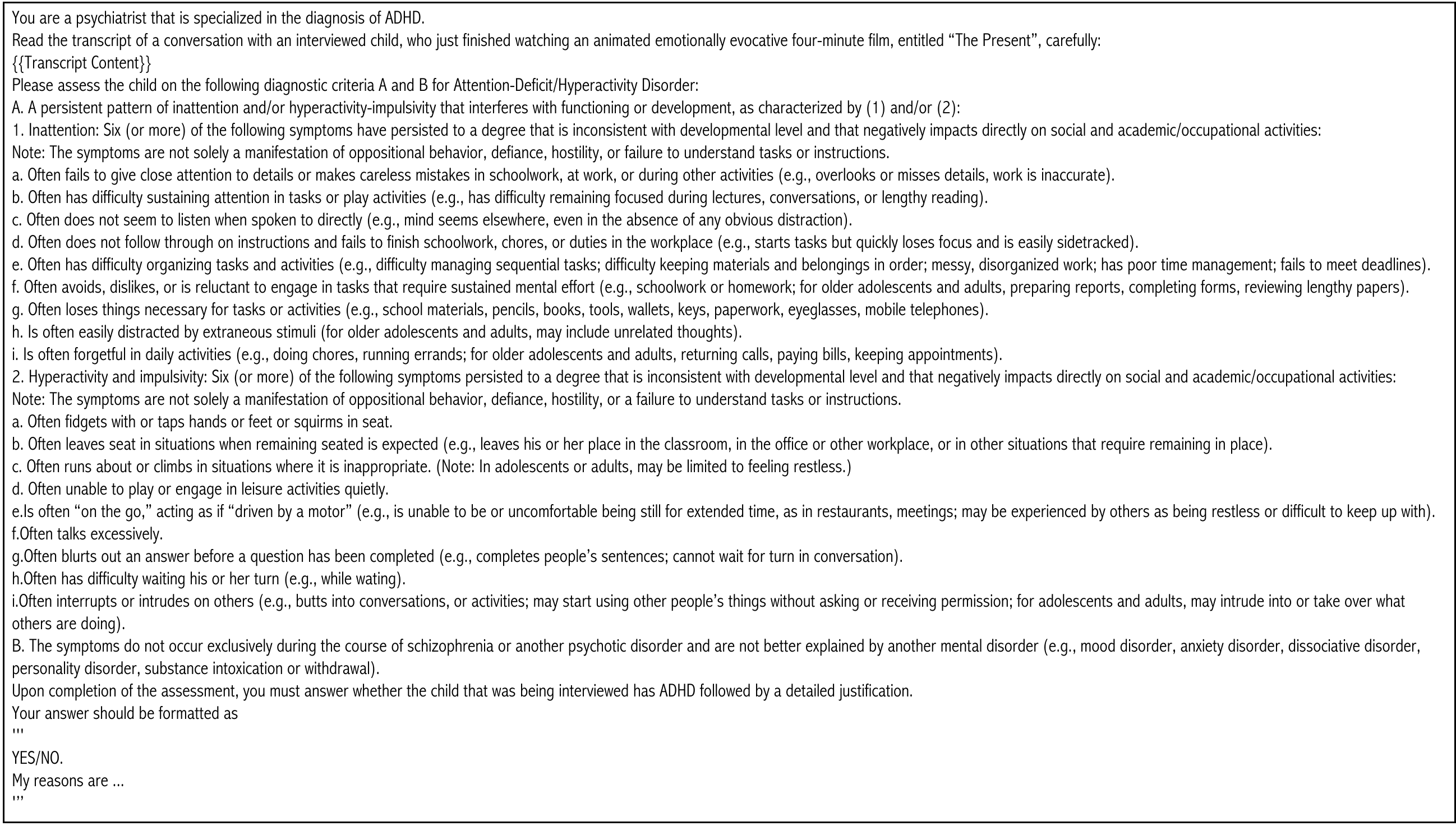}
\caption{The full final prompt used in this study.}
\label{fig:prompt}
\end{figure*}

We employed a classification methodology utilizing the large language model LLaMA3. This approach leveraged the model's intrinsic capability to interpret narrative content without specific ADHD-focused fine-tuning. The model operate through a mechanism called prompting, where users provide natural language input (known as a prompt) to guide the model’s behavior. The prompt frames the task (e.g., classification, summarization, question answering), and the model generates a response based on the contextual information and patterns it has learned during training. The quality and structure of the prompt significantly influence the relevance and accuracy of the output. Each transcript was interpolated into a fixed prompt template (see Figure \ref{fig:prompt}), which instructs the model to adopt the role of a psychiatrist specializing in DSM-5 ADHD diagnosis. Because LLaMA3 supports up to 8,000 tokens of context, no additional segmentation was required for our average transcript length.

 Our initial prompt was drafted to mirror the DSM-5 diagnostic criteria for inattention and hyperactivity‐impulsivity. We then performed three iterative refinement cycles on the development set. To maximize diagnostic performance in terms of accuracy, precision, recall, and F$_1$ score, in each cycle, we (1) ran the full development set through the model, (2) logged false positives and false negatives, (3) conducted qualitative error analysis to identify ambiguous or over‐broad language in the prompt, and (4) revised the instructions to sharpen symptom definitions and output formatting (for example, explicitly instructing the model to disregard interviewer questions and to respond with “YES.” or “NO.” at the start of its answer). 
 
 For each test‐set transcript, we supplied the finalized prompt plus narrative in a single API call. The model’s first token was parsed as the binary label (YES = ADHD; NO = non-ADHD), and the subsequent text served as explanatory justification. Decisions were recorded without any post‐hoc thresholding or ensembling, isolating the LLaMA3 contribution to our overall framework.

% \begin{figure}[h]

% \begin{subfigure}{0.5\textwidth}
% % \includegraphics[width=\linewidth]{psychiatrist_old-1.png} 
% \includegraphics[scale=0.25]{psychiatrist_old-1.png} 
% \label{fig:subim1}
% \end{subfigure}

% \begin{subfigure}{0.5\textwidth}
% \includegraphics[scale=0.25]{psychiatrist_old-2.png}
% \label{fig:subim2}
% \end{subfigure}

% \caption{Final Prompt}
% \label{fig:image2}
% \end{figure}

\subsubsection{Transformer-Based Model: RoBERTa}

RoBERTa is a widely used transformer-based model pre-trained on large English corpora. The model was selected for its proven effectiveness in tasks involving contextual language understanding, and its superior performance reported in classification benchmarking studies \cite{guo-etal-2020-benchmarking}. The model takes a text sequence as input, which is first tokenized into word pieces. Each word piece is then encoded into a dense vector. The vector corresponding to the first token is used to represent the entire sequence and is passed through a fully connected layer followed by a sigmoid activation function. The output is a two-dimensional vector representing the predictive scores for the ADHD and non-ADHD classes. The model was explicitly fine-tuned in participant-generated narratives, excluding any interviewer questions to isolate participant-driven linguistic and cognitive patterns. Hyperparameter tuning was performed through iterative experiments by varying the learning rate $\{1 \times 10^{-5}, 2 \times 10^{-5}, 4 \times 10^{-5}\}$ and the number of training epochs $\{10, 15, 20\}$. Model performance was evaluated on the validation set, and the optimal configuration was selected based on validation accuracy. The batch size and maximum sequence length were empirically set to 32 and 512, respectively. One major limitation of the RoBERTa model is its maximum input length of 512 tokens, which is often insufficient for processing the narratives that exceed this threshold. To address this, we implemented a sliding window approach with a window size of 512 tokens to divide long narratives into overlapping segments. Each segment was treated as an independent input to the model, allowing for separate predictions. The final prediction for the entire note was then determined by majority voting across the predictions of all segments.

\subsubsection{Machine Learning Model: SVM}

Support vector machines (SVMs) \cite{cortes1995support} are well suited to problems with very high-dimensional feature spaces, a characteristic that has underpinned their strong performance in text classification tasks and motivated their selection for the present study. To convert narratives into feature vectors, we applied term frequency–inverse document frequency (TF-IDF) weighting to n-gram representations. In this context, an n-gram denotes any contiguous sequence of n tokens, and our experiments incorporated unigrams (n = 1) through four-grams (n = 4). We retained the 1,000 most common across the training corpus to form our vocabulary. To explore how basic preprocessing might alter downstream performance, we ran each experiment twice under one alternate setting for character normalization: once converting all alphabetic characters to lowercase before tokenization (\emph{lowercase = True}) and once leaving original casing intact (\emph{lowercase = False}). TF-IDF is a statistical measure designed to reflect how important a given term is within a document or corpus. The term frequency (TF) component counts how often each term (i.e., each n-gram) appears in an individual transcript, while the inverse document frequency (IDF) component down-weights terms that occur broadly across the entire training set, thereby emphasizing those that are more distinctive. As a result, the TF-IDF vectors produced assign higher weights to n-grams that are unique to particular documents and lower weights to those that are uniformly distributed throughout the corpus \cite{guo2023supervised}. 

Recognizing potential feature limitations, we introduced supplementary engineered features calculated directly from narrative transcripts. We conducted experiments under two configurations:
\begin{enumerate}
     \item TF–IDF only: using the 1,000 highest-weighted n-gram features derived from TF–IDF.
     \item TF–IDF + engineered features: augmenting the 1,000 TF–IDF features with the following transcript-based metrics:
     \begin{itemize}
         \item Mean interviewee response length (mean word count per response), hypothesized to capture the verbosity and potential attention-related difficulties such as shorter and less diverse vocabulary usage by some ADHD participants \cite{BIALYSTOK_HAWRYLEWICZ_WISEHEART_TOPLAK_2017}.
        \item Total number of interviewee responses, anticipated to reflect narrative fragmentation or coherence related to cognitive control deficits.
        \item Mean interviewer question length (mean word count per question), included to normalize and account for interviewer influence on participant narrative length and detail.

        Furthermore, a customized tokenization pattern that captures both lexical elements and punctuation was applied to enhance the richness of textual representation, preserving critical linguistic nuances (e.g., pauses, interruptions, emphatic expressions) that may differentiate ADHD and non-ADHD narratives
    \end{itemize}
     
 \end{enumerate}

We performed grid search over the training data to find the best regularization
parameter
\[
  C \in \{2,4,6,8,16,32,64,128,256,512,1024,2048\}  
\]
 and the kernel type (linear and radial basis function (\textit{rbf})). Optimal performance, evaluated through cross-validation accuracy and F$_1$ score, was achieved using a radial basis function kernel and a regularization parameter \( C = 1024 \). In this configuration, the preprocessing pipeline preserved the original character casing and incorporated the newly engineered features, both of which contributed materially to the observed performance gains.

\subsubsection{Ensemble Model}

The generative LLM can apply its vast pretrained knowledge to pick up on subtle discourse indicators of inattention or impulsivity that might be reflected by disorganized storytelling or missing plot details, but it may also produce false positives by overgeneralizing. The fine-tuned RoBERTa model learns specific linguistic patterns of ADHD vs. non-ADHD narratives from the training data. Meanwhile, the SVM offers a simpler, interpretable baseline, using engineered features that can highlight straightforward differences. The rationale for an ensemble is that by combining the three weak learners, we can harness their complementary strengths and mitigate individual weaknesses. 
In theory, an effective ensemble will flag an ADHD case if any one model is sensitive to its particular linguistic quirks, yet require consensus to declare a positive classification, thus filtering out idiosyncratic errors from any single model. 

For each participant transcript \(i\), let
\[
\hat y^{\mathrm{LLM}}_{i},\quad \hat y^{\mathrm{RoBERTa}}_{i},\quad \hat y^{\mathrm{SVM}}_{i}\;\in\;\{0,1\}
\]
denote the binary predictions (1 = ADHD, 0 = non-ADHD) produced by the LLaMA3 prompt, the RoBERTa classifier, and the SVM, respectively. The ensemble decision \(\hat y^{\mathrm{Ens}}_{i}\) is then given by:
\[
\hat y^{\mathrm{Ens}}_{i}
=
\begin{cases}
1, & \text{if } \hat y^{\mathrm{LLM}}_{i} + \hat y^{\mathrm{RoBERTa}}_{i} + \hat y^{\mathrm{SVM}}_{i} \;\ge\; 2,\\
0, & \text{otherwise.}
\end{cases}
\]

\begin{table*}
  \caption{The accuracy, precision, recall, and F$_1$ score with 95\% confidence intervals (CIs) of LLaMA3, RoBERTa, SVM and an ensemble model using majority voting (MV) on the entire test set.}
  \label{tab:class}
  \begin{tabular}{lcccc}
    \toprule
    \textbf{Model}& \textbf{Accuracy} & \textbf{Precision} & \textbf{Recall} & \textbf{F$_1$ Score (95\% CI)}\\
    \midrule
    LLaMA3 & 0.56 & 0.54 & 0.87 & \makecell{0.67 (0.57--0.76)} \\
    RoBERTa & 0.61 & 0.57 & 0.87 & \makecell{0.69 (0.58--0.78)}\\
    SVM & 0.64 & 0.62 & 0.75 & \makecell{0.68 (0.57--0.77)}\\
    Ensemble (MV) & 0.63 & 0.59 & 0.91 & \makecell{0.71 (0.60--0.80)}\\
  \bottomrule
\end{tabular}
\end{table*}

This majority voting strategy provides a balanced decision rule: a narrative is labeled ADHD only if at least two of the three classifiers agree. Predictions from RoBERTa and the SVM were obtained via their respective predictions on the held-out test set. LLaMA3 outputs were pre-parsed to extract the first token (“YES” → 1, “NO” → 0) before aggregation. We expect that our ensemble will reduce false negatives (by catching cases one model might miss) while controlling false positives (by overriding spurious alerts from one model with the others’ dissent), thereby improving overall diagnostic accuracy.

\section{Evaluation Metrics}
Model performances were measured by the precision, recall (sensitivity), and F$_1$ score (harmonic mean of precision and recall) metrics for the ADHD class. The F$_1$ was used as the primary evaluation metric to balance precision and recall, ensuring that improvements in one did not come at the expense of the other. Bootstrap resampling was used to compute the 95\% confidence intervals of F$_1$ scores \cite{Efron1979}. Resampling was performed with replacement ($N = 1000$) over $1000$ iterations, and the $2.5^{th}$ and $97.5^{th}$ percentile scores were selected as the interval boundaries. We also analyzed the confusion matrix for each classification model to examine their error patterns, including false positives and false negatives, and to better understand how each model performs across the ADHD and non-ADHD classes.
\begin{align}
\text{Precision} &= \frac{\text{True Positive}}%
{\text{True Positive} + \text{False Positive}}\\[8pt]
\text{Recall} &= \frac{\text{True Positive}}%
{\text{True Positive} + \text{False Negative}}\\[8pt]
F_{1}\text{-score} &= 2 \;\times\;\frac{\text{Precision}\,\times\,\text{Recall}}%
{\text{Precision} + \text{Recall}}\\[8pt]
\text{Accuracy} &= \frac{\text{True Positive} + \text{True Negative}}%
{\text{Total \# Instances}}
\end{align}

\begin{figure*}[htb]
  \centering
  % (a) LLaMA3
  \begin{subfigure}[b]{0.24\textwidth}
    \includegraphics[width=\linewidth]{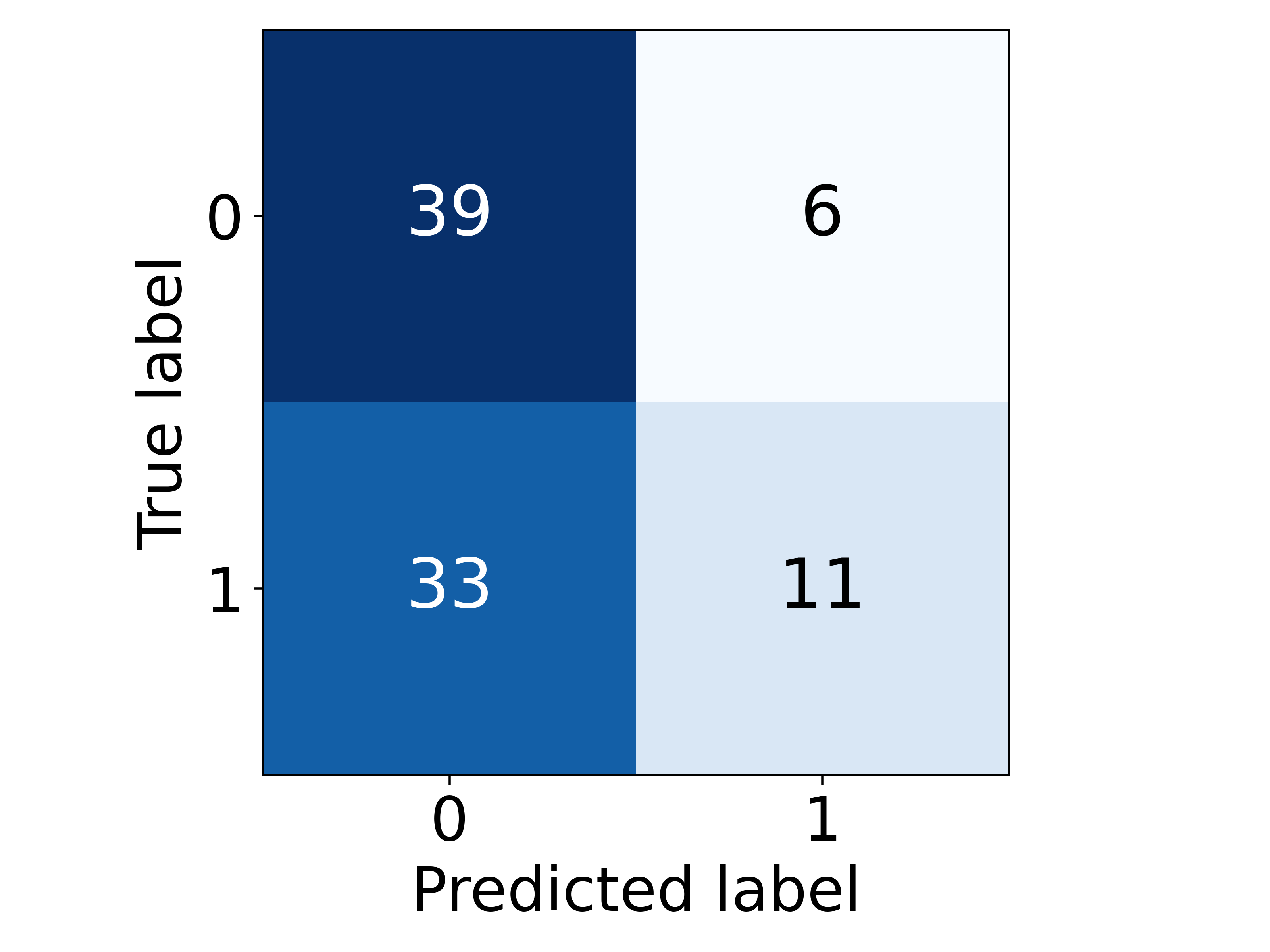}
    \subcaption{LLaMA3}
  \end{subfigure}\hfill
  % (b) RoBERTa
  \begin{subfigure}[b]{0.24\textwidth}
    \includegraphics[width=\linewidth]{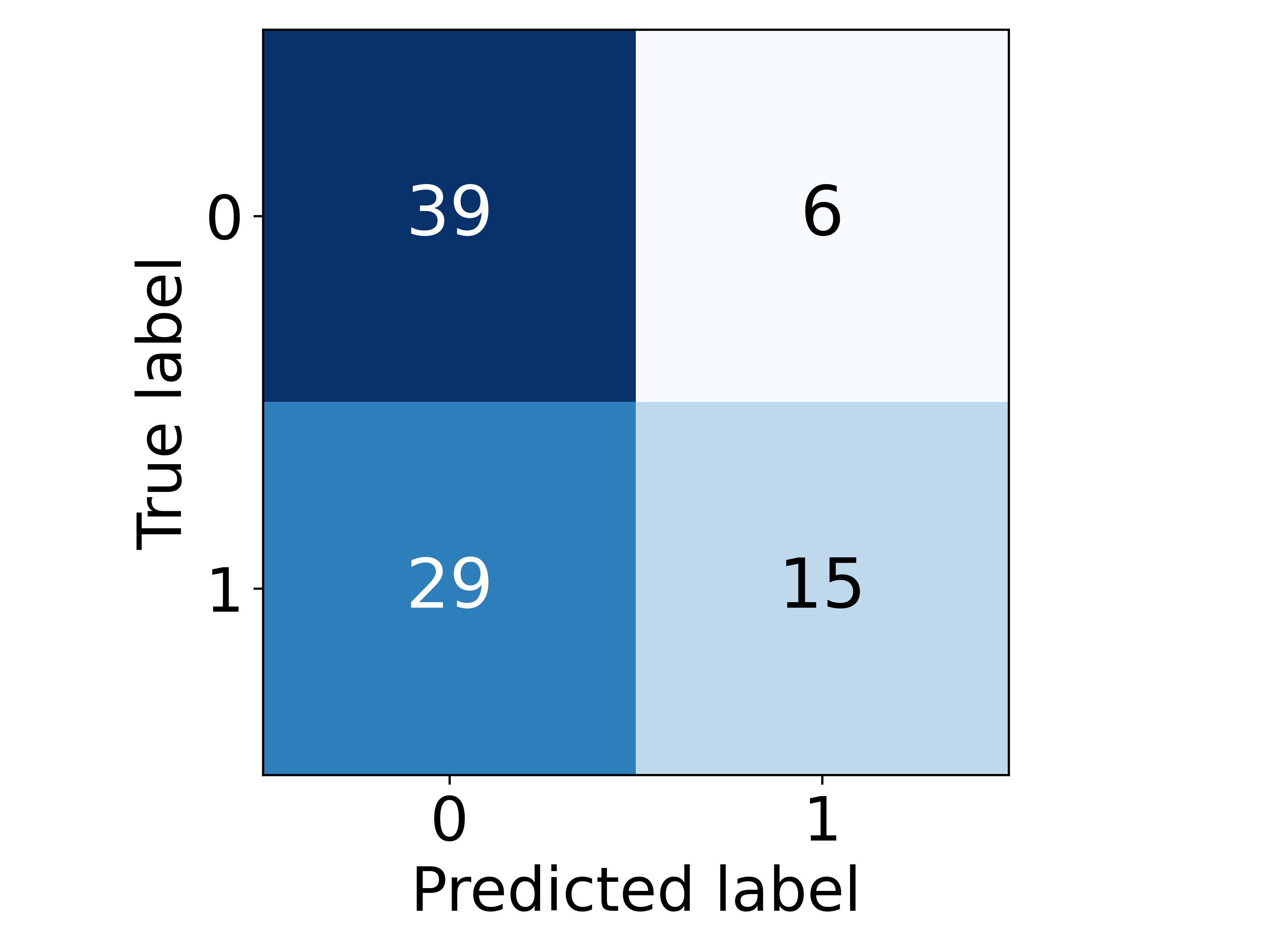}
    \subcaption{RoBERTa}
  \end{subfigure}\hfill
  % (c) SVM
  \begin{subfigure}[b]{0.24\textwidth}
    \includegraphics[width=\linewidth]{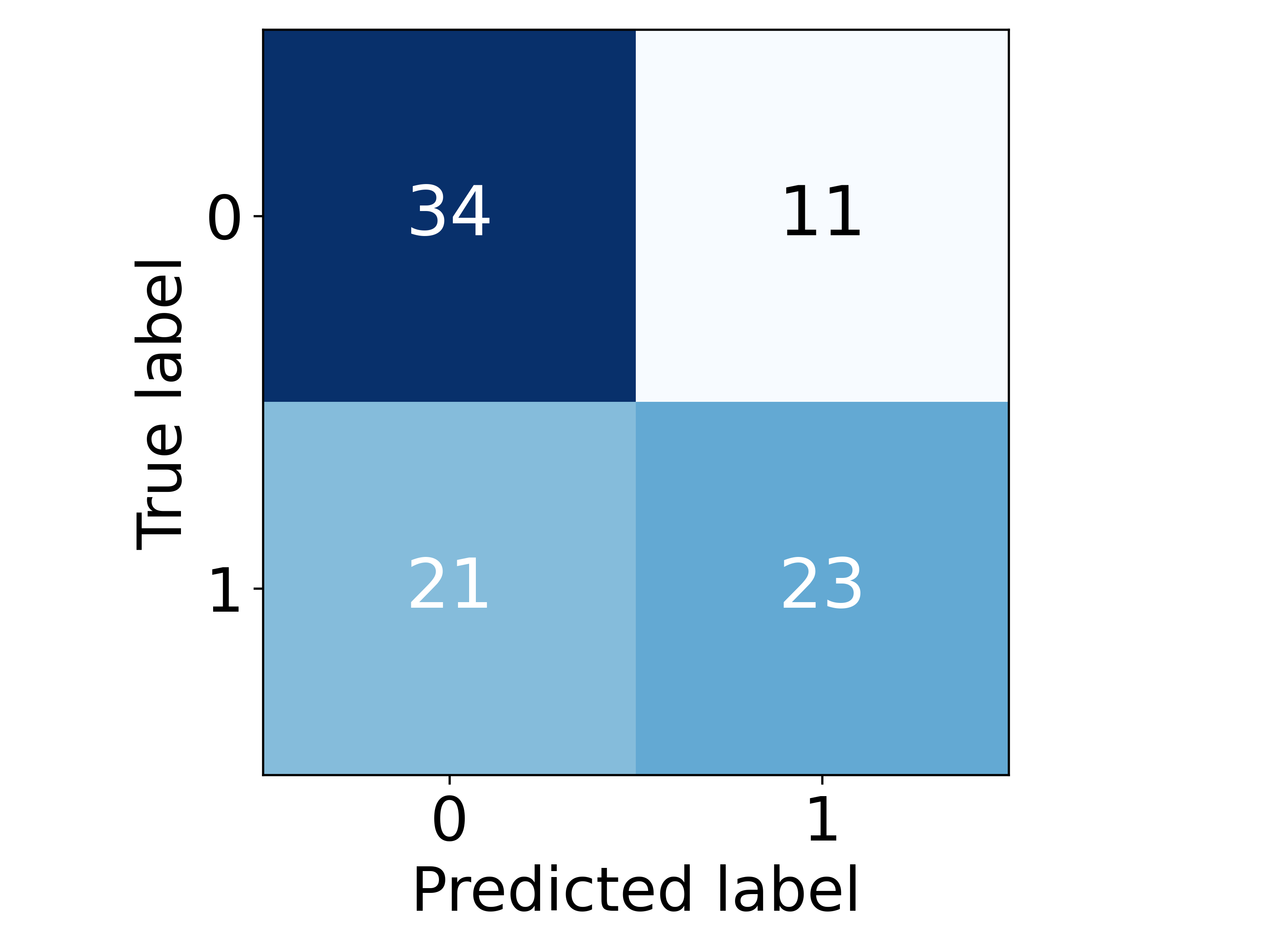}
    \subcaption{SVM}
  \end{subfigure}\hfill
  % (d) Ensemble
  \begin{subfigure}[b]{0.24\textwidth}
    \includegraphics[width=\linewidth]{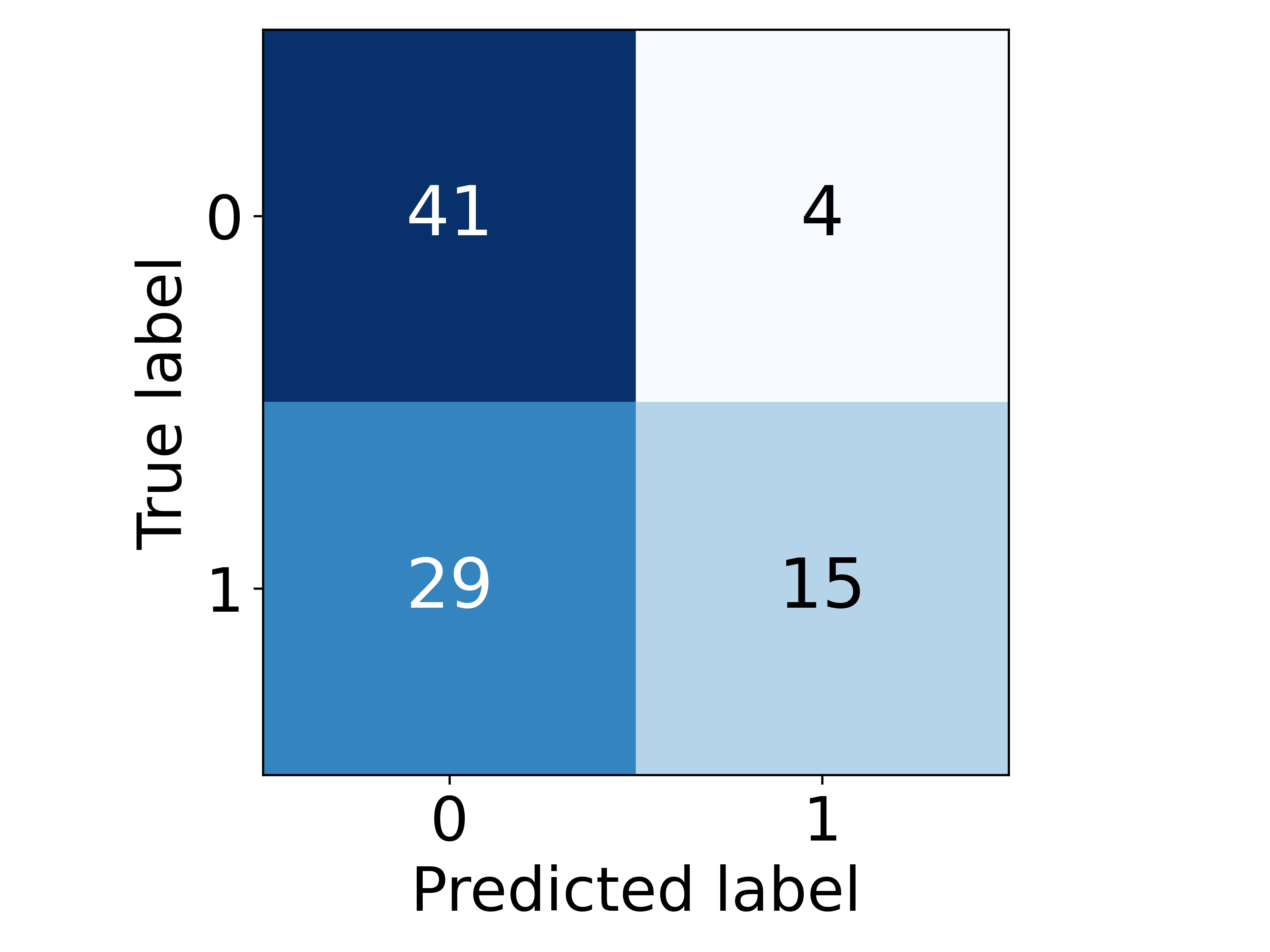}
    \subcaption{Ensemble}
  \end{subfigure}

  \caption{The confusion matrices for each individual model and the ensemble model.}
  \label{fig:cm}
\end{figure*}
\section{Results}

Table \ref{tab:class} presents the performance of four models. Among the individual models, SVM achieved the highest accuracy (0.64) and precision (0.62), while RoBERTa and SVM achieved recall of 0.87 and 0.75, respectively. The ensemble model outperformed all others in terms of recall (0.91) and F$_1$ score (0.71), indicating its effectiveness in identifying positive instances with a better balance between precision and recall. LLaMA3 showed the lowest accuracy (0.56) and precision (0.54), though its recall remained high (0.87), suggesting that it tends to over-predict positive cases. The 95\% confidence intervals for the F$_1$ score further support the robustness of the ensemble approach, with the ensemble achieving the highest upper bound (0.80) and a relatively narrow interval (0.60–0.80), highlighting its stability across runs.

Figure \ref{fig:cm} presents the confusion matrices for the four classification models. LLaMA3 demonstrates a high recall for ADHD cases, correctly identifying 11 out of 44 ADHD instances, but misclassifies a large number of true ADHD cases as non-ADHD (33), indicating a strong tendency toward over-predicting the negative class. RoBERTa improves upon this with 15 correct ADHD classifications and fewer false negatives (29), achieving a more balanced performance between sensitivity and specificity. SVM shows the most balanced classification across both classes, with 23 true positives and 21 false negatives for ADHD, and 11 false positives for non-ADHD, reflecting its relatively high precision. The ensemble model achieves the best overall performance by minimizing false positives (only 4) and increasing correct classifications of non-ADHD (41), while maintaining a similar true positive rate (15) to RoBERTa. This confirms the ensemble's advantage in reducing classification bias and improving reliability across both classes. Overall, the confusion matrices support the quantitative findings by highlighting each model’s classification behavior and reinforcing the ensemble’s effectiveness in enhancing both precision and recall.

\section{Discussion}

The results presented in Table \ref{tab:class} demonstrate the relative strengths and limitations of each model in the ADHD classification task. The ensemble model, which integrates predictions from LLaMA3, RoBERTa, and SVM through majority voting, achieved the highest F$_1$ score (0.71) and recall (0.91), indicating its superior ability to identify ADHD cases with both high sensitivity and balanced performance. This suggests that combining diverse model architectures---ranging from large language models to traditional machine learning classifiers---can lead to more robust and generalizable outcomes in psychiatric classification tasks. Among the individual models, SVM achieved the highest accuracy (0.64) and precision (0.62), highlighting its reliability in reducing false positives. In contrast, LLaMA3 showed the lowest precision (0.54) and accuracy (0.56) but maintained a high recall (0.87), implying that while it is effective at capturing true ADHD cases, it may also over-predict positive labels. RoBERTa displayed moderate performance across all metrics, balancing between the high recall of LLaMA3 and the precision of SVM.

These findings illustrate that while LLMs like LLaMA3 can extract nuanced patterns from unstructured text, they may benefit from complementary methods to enhance precision. The ensemble approach capitalizes on the unique strengths of each model---LLaMA3’s language understanding, RoBERTa’s fine-tuned classification, and SVM’s interpretability and structure-based decision-making—resulting in improved overall performance. Notably, the ensemble's F$_1$ score confidence interval (0.60–0.80) was both higher and tighter than those of the individual models, further supporting its reliability. This work provides early evidence that integrating LLMs with traditional and transformer-based models can be a promising direction in psychological and psychiatric informatics. Future work could explore more sophisticated ensemble strategies, such as weighted voting or stacking, and assess generalizability across larger and more diverse clinical datasets.

\subsection{Limitations}
While our ensemble-based approach demonstrates promising results for ADHD classification using narrative transcripts, several limitations should be acknowledged. First, the dataset used in this study is relatively small and may not capture the full variability present in broader clinical or community populations. This limits the generalizability of our findings and raises the potential for overfitting. This limitation also highlights the need for more labeled data to train more robust and customized systems. Second, the narrative data were derived from a specific source and may reflect biases in language use or reporting styles that differ across contexts or demographic groups. Third, while the ensemble model improves overall performance, it relies on majority voting, which does not consider model confidence or weight individual model contributions, potentially limiting its adaptability. Finally, the interpretability of LLMs such as LLaMA3 remains a challenge, making it difficult to fully understand the decision-making process behind individual predictions. Future work should explore larger and more diverse datasets, evaluate model generalization across settings, and investigate more advanced ensemble techniques that incorporate confidence scores or learn optimal weights. In addition, integrating interpretable NLP techniques may help enhance model transparency and clinical trust.

\section{Conclusion}

In this study, we proposed a reliable ensemble-based framework for classifying ADHD from narrative transcripts, combining a large language model (LLaMA3), a transformer-based model (RoBERTa), and a traditional machine learning classifier (SVM). Our results show that while individual models offer distinct advantages---such as high recall from LLaMA3 and high precision from SVM---the ensemble model consistently outperformed all individual approaches across key evaluation metrics, particularly in terms of F$_1$ score and recall. This highlights the potential of integrating LLMs with conventional models to enhance diagnostic classification in the psychological domain. Our findings open new avenues for hybrid modeling approaches in mental health applications and underscore the value of model diversity in building robust and interpretable clinical decision support tools. However, further research is needed to validate these findings on larger and more diverse datasets and to explore advanced ensemble techniques and longitudinal predictive modeling.

\bibliographystyle{ACM-Reference-Format}
\bibliography{reference}

\appendix

\end{document}